\documentclass[onecolumn]{dexforce}

\usepackage{tikz}
\usetikzlibrary{tikzmark, calc}
\usepackage{makecell}
\usepackage{float}

\algblockdefx[MainLoop]{MainLoop}{EndMainLoop}{\textbf{loop}}{\textbf{end loop}}

\makeatletter
\newcounter{subsubsubsection}[subsubsection]

\makeatother

\title{DexWorldModel: Causal Latent World Modeling towards Automated Learning of Embodied Tasks}

\author{
\begin{center}
    Yueci Deng \quad
    Guiliang Liu \quad
    Kui Jia \quad
    \\[12pt]
    DexForce AI
\end{center}
}

\begin{document}

\abstract{%
Deploying generative World-Action Models for manipulation is severely bottlenecked by redundant pixel-level reconstruction, $\mathcal{O}(T)$ memory scaling, and sequential inference latency. We introduce the Causal Latent World Model (CLWM), which employs DINOv3 features as generative targets to disentangle interaction semantics from visual noise, yielding highly robust domain generalization. To overcome memory scaling, CLWM features a Dual-State Test-Time Training (TTT) Memory that guarantees a strict $\mathcal{O}(1)$ footprint for long-horizon tasks. To overcome deployment latency, we propose Speculative Asynchronous Inference (SAI) to mask partial diffusion denoising behind physical execution, cutting blocking latency by about $50\%$. To scale robust policies, we present EmbodiChain, an online framework that establishes the Efficiency Law by injecting an infinite flow of physics-grounded trajectories during training. Extensive experiments validate that CLWM achieves state-of-the-art performance in complex dual-arm simulation and unprecedented zero-shot sim-to-real transfer on physical robots, outperforming baselines explicitly finetuned on real-world data.
}

\maketitle

\justifying
\section{Introduction}

Vision-Language-Action (VLA) models~\citep{zitkovich2023rt, kim2024openvla, team2024octo, black2024pi_0, black2025pi_, wen2025tinyvla, bjorck2025gr00t, team2026gigabrain} have shown remarkable promise in enabling robots to follow language instructions in complex environments. However, standard feed-forward VLAs entangle high-dimensional visual understanding, physical dynamics, and low-dimensional motor control within a single representation space, fundamentally restricting their capacity for explicit causal reasoning~\citep{lecun2022path}. To address this, World Action Models (WAMs)~\citep{li2025unified, zhu2025unified, bi2025motus, li2026causal, ye2026world} jointly perform forward frame prediction and policy execution. By empowering robots to "imagine" future states, this generative paradigm grounds action inference in physical causality rather than spurious visual correlations.

Despite their progress in closed-loop control~\citep{li2026causal, ye2026world}, existing WAMs predominantly model future states directly in the pixel or VAE latent space~\citep{wan2025wan}. This inevitably couples state transition learning with the generation of redundant visual details, e.g., illumination variations and cluttered backgrounds. Expending substantial model capacity on reconstructing task-irrelevant pixels hinders the extraction of interaction-centric features~\citep{lyu2026lda} and severely limits generalization during sim-to-real or real-to-real domain shifts. To overcome this representational bottleneck, we propose the Causal Latent World Model (CLWM), which employs DINOv3 latent features as generative targets instead of low-level textural tokens. Since the structured DINOv3 space is naturally robust to visual noise and background variations~\citep{oquab2023dinov2, simeoni2025dinov3}, formulating sequence generation within this semantic space effectively bypasses the computational burden of pixel reconstruction. This disentanglement dedicates the model's capacity strictly to the temporal evolution of interaction semantics, preventing overfitting to superficial textures and yielding significantly more robust domain generalization.

Equipping WAMs with long-term memory for long-horizon manipulation introduces severe memory and computational bottlenecks. In standard causal world models~\citep{li2026causal, ye2026world}, the size of the KV cache~\citep{pope2023efficiently} scales linearly with the autoregressive generation steps. This unbounded accumulation leads to prohibitive memory footprints and escalating inference latency during prolonged physical interactions. While recent VLA models attempt to mitigate this via external memory banks~\citep{shi2025memoryvla} or text-space summarization~\citep{torne2026mem}, these heuristics suffer from lossy cross-modal compression and fail to resolve the underlying architectural inefficiency. To overcome this computational limitation, we replace the conventional KV cache in CLWM with a Test-Time Training (TTT) layer~\citep{sun2024learning}. Building upon the efficacy of TTT in long-context language modeling~\citep{sun2024learning, behrouz2024titans, behrouz2025atlas} and video generation~\citep{JMLR:v26:24-0439, dalal2025one}, our approach parameterizes all historical video and action observations into an embedded, test-time updatable multi-layer perceptron (MLP). By internalizing the context into dynamic model weights, CLWM maintains a strictly constant $\mathcal{O}(1)$ memory footprint regardless of the physical interaction trajectory length. This design fundamentally eliminates the sequence-scaling memory burden, unlocking highly efficient, unconstrained long-term memory for complex robotic manipulation.

Furthermore, high-frequency closed-loop control is imperative for robust physical deployment. Conventional VLA policies are fundamentally reactive, strictly conditioning action inference on real-time sensor inputs~\citep{kim2024openvla, black2024pi_0}. This dependency induces a severe sequential latency bottleneck: the model remains idle during physical execution, awaiting the next sensory observation before initiating subsequent predictions. By contrast, CLWM acts as a generative world model, concurrently synthesizing future action chunks and their corresponding visual representations. Leveraging this forward-predictive capacity, we propose a Speculative Asynchronous Inference (SAI) strategy. SAI decouples inference from real-time observation by utilizing self-generated future video features as surrogate conditions. While the robot physically executes the current action chunk, CLWM proactively performs early-stage diffusion "pre-denoising" for the next time step. Once the physical ground-truth returns, the model instantly calibrates its hidden state and executes only the minimal remaining fine-grained denoising steps. This asynchronous overlap between expensive diffusion sampling and physical execution drastically elevates the end-to-end control frequency. Empirically, within the RoboTwin simulator~\citep{chen2025robotwin}, SAI reduces per-chunk inference latency by up to $50\%$ compared to the state-of-the-art autoregressive baseline, Lingbot VA~\citep{li2026causal}.

Beyond architectural innovations, scaling robot learning requires robust training paradigms. While utilizing massive public datasets, such as RoboMind2~\citep{wu2024robomind, hou2025robomind}, AgiBot World~\citep{bu2025agibot} and InternData-A1~\citep{contributors2025internroboticsrepo}, is crucial for pre-training fundamental physical priors, we introduce EmbodiChain~\citep{EmbodiChain} to revolutionize the post-training phase. EmbodiChain establishes the \textit{Efficiency Law} of embodied intelligence via continuous Online Data Streaming (ODS). Rather than relying on finite, static datasets, it generates a high-throughput stream of physics-grounded, functionally diverse trajectories and injects them directly into the optimizer. This continuous influx of novel experiences prevents homogenization and bridges the sim-to-real gap autonomously.

We comprehensively evaluate CLWM across the challenging RoboTwin simulated benchmark~\citep{chen2025robotwin} and complex physical robot deployments. Extensive experiments demonstrate that CLWM establishes a new state-of-the-art in dual-arm manipulation. Crucially, fueled by our architecture and EmbodiChain's online data streaming, CLWM achieves dominating zero-shot sim-to-real transfer on physical hardware, decisively outperforming established baselines even when the latter are explicitly finetuned on real-world human demonstrations.

Overall, our main contributions are as follows:
\begin{enumerate}
    \item \textbf{Causal Latent World Model (CLWM)}: We propose a causal latent world model that employs DINOv3 latent features as generative targets, effectively disentangling interaction semantics from redundant pixel reconstruction for superior domain generalization.
    \item \textbf{Constant-Memory Allocation via TTT}: We innovatively replace the conventional KV cache with a Dual-State Test-Time Training (TTT) Memory, achieving a strict $\mathcal{O}(1)$ memory footprint that unlocks unconstrained reasoning for long-horizon manipulation.
    \item \textbf{Speculative Asynchronous Inference (SAI)}: Leveraging the forward-predictive capacity of CLWM, we introduce an asynchronous inference strategy that masks diffusion pre-denoising behind physical execution, reducing blocking latency by about $50\%$.
    \item \textbf{EmbodiChain Generative and Online Training Paradigm}: We operationalize the Efficiency Law via an online, closed-loop simulation framework (ODS), enabling CLWM to achieve unprecedented zero-shot sim-to-real transfer capabilities on physical robots.
\end{enumerate}

\section{Preliminaries}

\subsection{Vision-Language-Action Models}

Formally, we model robotic manipulation as a Partially Observable Markov Decision Process~\citep{lauri2022partially}. At any given time step $t$, the robot receives a high-dimensional visual observation $o_t \in \mathbb{R}^{H \times W \times 3}$ and operates under a task-specifying natural language instruction $l$. Conventional VLA models~\citep{zitkovich2023rt, kim2024openvla, team2024octo} instantiate a feed-forward policy network $\pi_\theta$ to map the accumulated cross-modal context directly to the low-dimensional action space $\mathcal{A}$. The objective is to predict a sequence of future action chunks $a_{t:t+K-1}$, where $K$ is the action chunk length:
\begin{equation}
a_{t:t+K-1} \sim \pi_\theta(\cdot \mid o_{\le t}, l)
\end{equation}

While modern VLAs, e.g., $\pi_0$~\citep{black2024pi_0}, increasingly parameterize this continuous action distribution using expressive generative frameworks, they remain fundamentally reactive systems. By implicitly mapping high-dimensional scene understanding directly to motor commands, VLAs bypass explicit forward dynamics modeling. This representational entanglement inherently restricts their capacity for physical causal reasoning and proactive long-horizon planning~\citep{lecun2022path}.

\subsection{World Action Models}

To overcome the reactive limitations of standard VLAs and endow robots with the ability to "imagine" future states, recent studies introduce World Action Models (WAMs)~\citep{li2026causal, ye2026world}. Rather than directly optimizing an entangled policy distribution, WAMs recast robotic control from a pure mapping problem into a causal autoregressive generation paradigm.Given a unified multimodal sequence $s_{<t} = \{(o_1, a_1), \dots, (o_{t-1}, a_{t-1})\}$, WAMs factorize the world modeling process into two sequential probabilistic stages:

\noindent\textbf{(1) Forward Visual Dynamics}: The model first anticipates how the visual world will evolve by predicting the future state conditioned on the historical context:
\begin{equation}
\hat{o}_{t+1} \sim p_\theta(\cdot \mid o_{\le t}, a_{<t}, l)
\end{equation}

\noindent\textbf{(2) Inverse Dynamics}: Subsequently, an inverse dynamics model decodes the requisite motor commands necessary to transition the environment to the predicted future state:
\begin{equation}
a_t \sim g_\psi(\cdot \mid o_{\le t}, a_{<t}, \hat{o}_{t+1}, l)
\end{equation}

By sequentially predicting future states and inferring actions, this two-stage formulation effectively grounds policy execution in explicit physical causality. However, directly computing and sampling from these high-dimensional continuous distributions, especially the pixel-space $o_{t+1}$, is mathematically intractable, necessitating the adoption of advanced generative frameworks, such as conditional flow matching.

\subsection{Conditional Flow Matching}

To synthesize the high-dimensional continuous distributions formulated in WAMs, recent architectures typically employ Conditional Flow Matching (CFM)~\citep{lipman2022flow, davtyan2023efficient, ni2023conditional, wan2025wan} as the underlying generative backbone. CFM is a continuous-time generative modeling framework that learns to smoothly transform a simple, tractable prior noise distribution into a complex target data distribution through an Ordinary Differential Equation (ODE) flow.Let the target generative state be $x$ (which, in the two-stage WAM paradigm, can represent either the predicted future visual frame $o_{t+1}$ or the action command $a_t$), and let the source noise be $\epsilon \sim \mathcal{N}(0, I)$. CFM defines a time-dependent conditional vector field $v_\phi(x^{(s)}, s \mid c)$ to describe the instantaneous velocity of particles flowing from the noise $\epsilon$ to the true data $x$:
\begin{equation}
\frac{dx^{(s)}}{ds} = v_\phi(x^{(s)}, s \mid c), \quad x^{(0)} = \epsilon \sim \mathcal{N}(0, I),
\end{equation}
where $s \in [0, 1]$ denotes the continuous flow time, and $c$ represents the conditioning context required for the generation process, i.e., the historical observation sequence and language instruction $c = (o_{\le t}, a_{<t}, l)$.

Following the optimal transport path formulation~\citep{mccann1997convexity}, the linear interpolation between the source noise and the target data is defined as $x^{(s)} = (1 - s)\epsilon + s\cdot x$. Taking the derivative of this path with respect to time $s$ yields a constant target velocity $\dot{x}^{(s)} = x - \epsilon$. Consequently, the neural network is optimized by minimizing the following vector field regression objective:
\begin{equation}
\mathcal{L}_{\text{CFM}} = \mathbb{E}{s, \epsilon, x, c} \left[ || v_\phi(x^{(s)}, s \mid c) - \dot{x}^{(s)} ||^2 \right]
\end{equation}

During inference, starting from the initial random noise $\epsilon$, an ODE solver, e.g., the Euler method, iteratively integrates the learned conditional vector field $v_\phi$ to progressively denoise and sample the predicted future states or actions.

While CFM provides a mathematically rigorous and highly efficient generative engine for WAMs, applying it directly within the traditional WAM paradigm exposes two fatal architectural bottlenecks. First, defining the target state $x = o_{t+1}$ in the raw pixel space forces the flow matching process to waste substantial computational capacity fitting task-irrelevant, high-frequency textures, yielding no actionable signal for downstream robotic control. Second, to process the continuously accumulating historical condition $c$, autoregressive generation models must maintain a KV cache that scales linearly $\mathcal{O}(T)$, inevitably leading to severe memory exhaustion during long-horizon manipulation. We fundamentally address these two bottlenecks by proposing the Causal Latent World Model in the subsequent section.
\section{Causal Latent World Model}

\begin{figure}[!t]
    \centering
    \includegraphics[width=\linewidth]{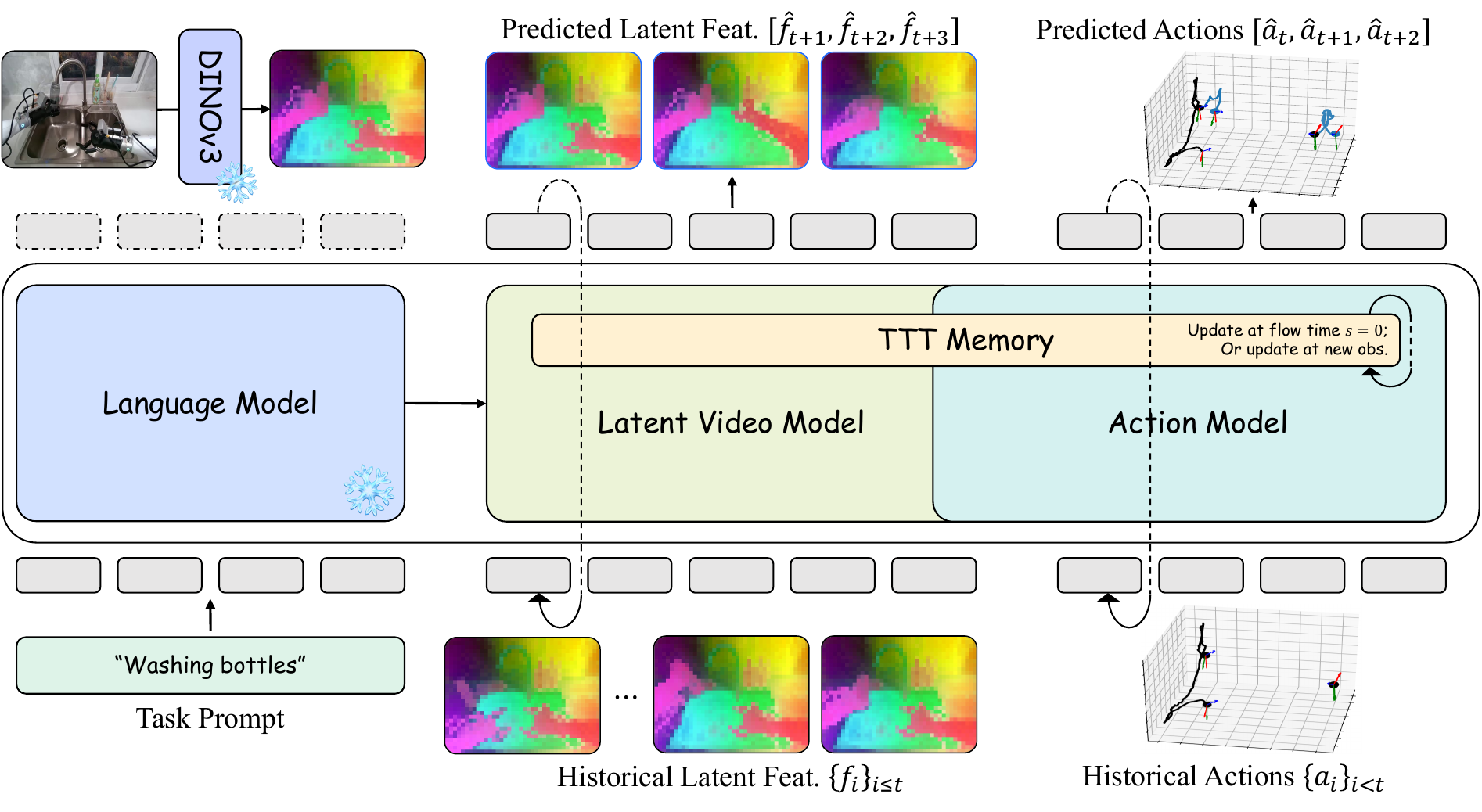}
    \caption{Overview of the Causal Latent World Model (CLWM). CLWM employs a Mixture of Transformers (MoT) architecture that unifies a latent video model and an action model. To maintain historical context across interleaved latent frame and action tokens, a shared Test-Time Training (TTT) memory module dynamically updates its hidden states at flow time $s=0$ (working memory for action generation) or arriving new observations (long-term memory). With the latent video features obtained by a frozen DINOv3 model, CLWM performs autoregressive generation: the latent video model first predicts future features via flow matching, which subsequently conditions the action model to decode the corresponding action chunks.}
    \label{fig:pipeline}
\end{figure}

To overcome the representational redundancy and memory exhaustion inherent in conventional World Action Models, we propose the Causal Latent World Model (CLWM). As illustrated in Fig.~\ref{fig:pipeline}, CLWM fundamentally redefines the generative state space through a world action model and replaces the explicit historical conditioning with an implicitly updated Test-Time Training (TTT) memory module. Furthermore, leveraging the forward-predictive capacity of world models, we introduce a Speculative Asynchronous Inference (SAI) strategy to eliminate sequential latency bottlenecks during physical deployment.

\subsection{Autoregressive Generation of Latent Video Features and Actions}

In standard WAMs, predicting future states, such as visual observations, within the raw pixel space or a texture-level VAE latent space forces the generation process to waste substantial computational capacity on reconstructing task-irrelevant textures, yielding minimal actionable signals for downstream control. To strictly dedicate the generative backbone to the temporal evolution of interaction semantics, CLWM eschews low-level pixel modeling. Instead, we employ the pre-trained DINOv3 base model~\citep{simeoni2025dinov3} as a robust feature extractor to derive high-level semantic representations:
\begin{equation}
f_t = \Phi_{\text{DINO}}(o_t) \in \mathbb{R}^{C \times H' \times W'},
\end{equation}
where $H' = H/P$ and $W' = W/P$ denote the spatial dimensions of the latent features, with $P=16$ being the default patch size of the DINOv3 base model.

To process these features, CLWM adopts a highly parameter-efficient Mixture of Transformers (MoT) paradigm. Specifically, the Latent Video Model ($\phi_{\text{vid}}$) and the Action Model ($\phi_{\text{act}}$) share their core transformer blocks, initialized from Wan2.2-5B~\citep{wan2025wan}, to learn universal environmental dynamics. Domain independence is strictly confined to the flow timestep embeddings and the randomly initialized linear input/output projection layers:
\begin{equation}
\phi_{\text{vid}} = \phi_{\text{vid}}^{\text{out}} \circ \phi_{\text{share}} \circ \phi_{\text{vid}}^{\text{in}}; \quad \phi_{\text{act}} = \phi_{\text{act}}^{\text{out}} \circ \phi_{\text{share}} \circ \phi_{\text{act}}^{\text{in}}
\end{equation}

This elegant parameter-sharing strategy inherently enforces deep cross-modal alignment while enabling an Autoregressive Flow Matching process that sequentially predicts future latent video states and actions.

\noindent\textbf{Stage 1: Latent Video Flow Matching.} Given the historical memory context $h_{\le t}$ and language instruction $l$, the Latent Video Model utilizes its domain-specific projections and the shared backbone to estimate the vector field required to denoise random noise $\epsilon_{\text{vid}}$ into the structured future latent feature $x_{\text{vid}} = f_{t+1}$. The regression objective is formulated as:
\begin{equation}
\mathcal{L}_{\text{video}} = \mathbb{E}_{s, \epsilon_{\text{vid}}, f_{t+1}, h_t, l} \left[ \left| v_{\phi_{\text{vid}}}(f_{t+1}^{(s)}, s \mid h_{\le t}, l) - \dot{f}{t+1}^{(s)} \right|^2 \right]
\end{equation}
where $f_{t+1}^{(s)} = (1-s) \epsilon_{\text{vid}} + s\cdot f_{t+1}$ represents the state at flow time $s \in [0, 1]$ with noise $\epsilon_{\text{vid}} \sim \mathcal{N}(0, I)$, and $\dot{f}_{t+1}^{(s)} = f_{t+1} - \epsilon_{\text{vid}}$ is the ground-truth flow velocity along the interpolation path.

\noindent\textbf{Stage 2: Action Flow Matching.} Subsequently, the Action Model decodes the corresponding action chunks $a_t = \{a_{t, 1}, a_{t, 2}, \dots, a_{t, \tau}\}$, where $\tau$ denotes the action chunk size, representing the temporal frequency ratio between action tokens and visual latent frames (empirically set to $\tau=16$ in our experiments). The conditional vector field explicitly attends to the historical context $h_t$, the language $l$, and the predicted future semantics $\hat{f}_{t+1}$ inferred from Stage 1.

To enhance the robustness of the Action Model against imperfect visual histories during simulation testing, we leverage a history augmentation strategy~\citep{li2026causal}. During training, we inject Gaussian noise of varying scales into the historical latent features $f_{\le t}$ with a probability of $p=0.5$:
\begin{equation}\tilde{f}_{\le t} =
\begin{cases}
(1 - s_{\text{aug}})\epsilon + s_{\text{aug}}\cdot f_{\le t}, & p = 0.5,\ s_{\text{aug}} \in [0.5, 1],\ \epsilon \sim \mathcal{N}(0, I) \\
f_{\le t}, & 1 - p = 0.5\end{cases}\label{eq:augment_history}
\end{equation}
By replacing the clean history with this augmented context $\tilde{h}_{\le t} = (\tilde{f}_{\le t}, a_{<t})$, we encourage the Action Model to learn how to deduce accurate control signals from noisy trajectories. This enables the generation of high-precision actions even when relying on partially denoised latent video states during deployment. Setting the generative target to $x_{\text{act}} = a_t$, the objective becomes:
\begin{equation}
\mathcal{L}_{\text{action}} = \mathbb{E}_{s, \epsilon_{\text{act}}, a_t, \tilde{h}_t, l, \tilde{f}_{t+1}} \left[ \left| v_{\phi_{\text{act}}}(a_t^{(s)}, s \mid \tilde{h}_{\le t}, l, \tilde{f}_{t+1}) - \dot{a}_{t}^{(s)} \right|^2 \right]
\end{equation}
where $a_{t}^{(s)} = (1-s)\epsilon_{\text{act}} + s\cdot a_t$ and $\dot{a}_t^{(s)} = a_t - \epsilon_{\text{act}}$.

This autoregressive generation effectively grounds action decoding in physical forward anticipation, while alleviating the computational burden of low-level texture reconstruction to enforce modeling at the level of semantic dynamics.

\subsection{Constant-Memory Autoregression via Test-Time Training}

\begin{figure}
    \centering
    \includegraphics[width=\linewidth]{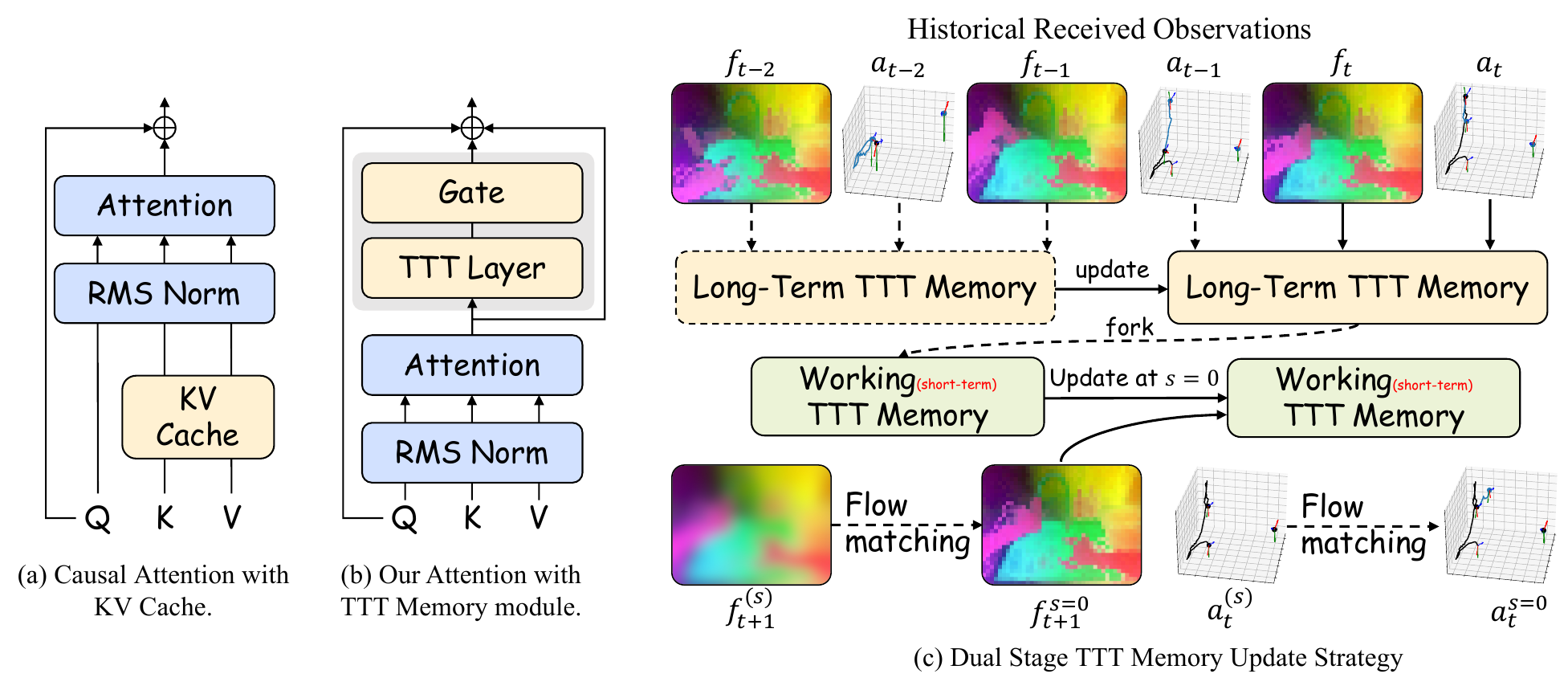}
    \caption{Architecture of the TTT Memory Module. (a) Standard causal attention relies on a KV cache to maintain historical context. (b) Our architecture replaces the KV cache with a Test-Time Training (TTT) Layer. (c) The Dual-State TTT Memory Update Strategy. We maintain a \textit{Long-Term TTT Memory} updated exclusively by real historical observations. For each generation step, a \textit{Working (Short-Term) TTT Memory} is forked from the long-term weights to condition the latent video generation. It remains frozen during the ODE integration but is immediately updated at flow time $s=0$ using the predicted latent state.}
    \label{fig:TTT-Memory}
\end{figure}

In conventional autoregressive world models, generating future states or actions conditioned on a continuously accumulating history requires the maintenance of an explicit KV cache, as Fig.~\ref{fig:TTT-Memory}(a). For standard Transformer architectures, the spatial complexity of this KV cache scales linearly, $\mathcal{O}(T)$, with the sequence length. In the context of embodied AI, where agents must continuously perceive and act over extended temporal horizons, this unbounded memory accumulation inevitably leads to severe memory exhaustion and unacceptable inference latency.

To fundamentally eliminate this sequence-scaling memory barrier, we draw inspiration from recent breakthroughs in Test-Time Training (TTT) for sequence modeling~\citep{sun2024learning, JMLR:v26:24-0439} and propose the \textbf{TTT Memory Module}. Instead of storing an ever-growing sequence of past multimodal tokens, the TTT Memory internalizes the entire historical context into the dynamic weights of an implicitly updatable neural layer, as Fig.~\ref{fig:TTT-Memory}(b).

\noindent\textbf{Definition of the TTT Layer.} Following \cite{sun2024learning}, we instantiate the core component of our module as a \textbf{TTT-MLP}. For a given input, e.g., the latent frame or action tokens $z_t\in \mathbb{R}^{L\times D}$ at timestep $t$, the self-supervised reconstruction task is parameterized by learnable low-rank projections $\theta_K, \theta_V$, and $\theta_Q$, analogous to the Key, Value, and Query weight matrices in standard self-attention. The self-supervised loss $\ell_{\text{self}}$ optimizes the hidden state weights $\mathcal{W}$ to reconstruct a projected target $\theta_V z_t$ from a projected input $\theta_K z_t$:
\begin{equation}
\ell_{\text{self}}(\mathcal{W}; z_t) = \left|| f(\theta_K z_t; \mathcal{W}) - \theta_V z_t \right||^2
\end{equation}

Here, the inner-loop model $f$ is formulated as a residual block wrapping a two-layer multi-layer perceptron: $f_{TTT_{mlp}}(x; \mathcal{W}) = x + \text{LN}(\text{MLP}(x; \mathcal{W}))$, where LN denotes Layer Normalization and the MLP utilizes a GELU activation with a $4\times$ expansion factor. Once the weights are updated to $\mathcal{W}_t$ via gradient descent, the output hidden state is extracted utilizing the query projection:
\begin{equation}
l_t = f_{TTT_{mlp}}(\theta_Q z_t; \mathcal{W}_t)
\end{equation}

Additionally, naively inserting TTT layers into a pre-trained network would dramatically worsen its predictions at the beginning of fine-tuning, we gate TTT with a learned vector $\alpha\in \mathbb{R}^D$ following standard practice, as \cite{JMLR:v26:24-0439}:

\begin{equation}
    f_{TTT}(z_t; \mathcal{W}_t) = tanh(\alpha) \otimes f_{TTT_{mlp}}(\theta_Q z_t; \mathcal{W}_t) + z_t,
\end{equation}
where we initialize all values in $\alpha$ to $0.1$ at the
beginning of fine-tuning.

Crucially, these projection matrices ($\theta_K, \theta_V, \theta_Q$) are optimized in the outer loop during the standard training phase, while the network weights $\mathcal{W}$ are updated dynamically at test time (the inner loop).

\noindent\textbf{Dual-State TTT Memory Update Strategy.} To accommodate the distinct temporal requirements of our autoregressive flow matching formulation, specifically, maintaining static conditions during ODE integration while updating context between cascaded stages, we encapsulate the TTT-MLP within a Dual-State TTT Memory architecture comprising a \textit{Long-Term TTT Memory} and a \textit{Working TTT Memory}, as Fig.~\ref{fig:TTT-Memory}(c).

\noindent\textbf{1. Long-Term TTT Memory (The Anchor).}
We maintain a persistent set of parameters, denoted as $\mathcal{W}^{\text{long}}$, dedicated to anchoring the true physical history. This long-term memory is updated exclusively when new ground-truth observations and executed actions are received from the physical environment. Given the newly received historical state $h_t = [f_t, a_{t-1}]$, the long-term memory performs an online gradient update:
\begin{equation}
\mathcal{W}_t^{\text{long}} = \mathcal{W}_{t-1}^{\text{long}} - \eta \nabla_{\mathcal{W}} \ell_{\text{self}}(\mathcal{W}_{t-1}^{\text{long}}; h_t)
\end{equation}
where $\eta$ is the test-time learning rate. This operation strictly confines the true environmental causality within $\mathcal{W}_t^{\text{long}}$, maintaining a strict $\mathcal{O}(1)$ spatial complexity regardless of the trajectory length.

\noindent\textbf{2. Working TTT Memory (The Fork).} During the generative phase at step $t$, we must sequentially predict the future latent video $f_{t+1}$ and the action $a_t$. To avoid corrupting the ground-truth history, we \textit{fork} (clone) a transient set of parameters from the long-term memory:
\begin{equation}
\mathcal{W}_t^{\text{work}} \leftarrow \mathcal{W}_t^{\text{long}}
\end{equation}

\noindent In \textbf{Stage 1} (Latent Video Generation), the Working TTT Memory extracts the conditioning hidden state $l_t = f_{TTT}(f^{(s)}_{t+1}; \mathcal{W}_t^{\text{work}})$. Importantly, to ensure mathematical stability during the continuous-time ODE integration of the flow matching process, the working weights $\mathcal{W}_t^{\text{work}}$ remain strictly frozen across all intermediate flow steps $s \in (0, 1]$.

\noindent\textbf{3. Intermediate Update at $s=0$.} Once the latent video prediction concludes at flow time $s=0$, yielding the predicted future feature $\hat{f}_{t+1}$, the Working Memory undergoes an instantaneous intermediate update. It absorbs this predicted future to provide accurate contextual conditioning for the subsequent action decoding:
\begin{equation}
{\mathcal{W}_t^{\text{work}}}' \leftarrow \mathcal{W}_t^{\text{work}} - \eta \nabla_{\mathcal{W}} \ell_{\text{self}}(\mathcal{W}_t^{\text{work}}; \hat{f}_{t+1})
\end{equation}

\noindent In \textbf{Stage 2} (Action Generation), the Action Model utilizes the newly updated hidden state $l_t' = f_{TTT}(a_t^{(s)}; {\mathcal{W}_t^{\text{work}}}')$ to guide the vector field.

By strictly isolating true physical observations within the Long-Term Memory and confining predictive, transient context to the Working Memory, this dual-state strategy elegantly prevents causal confusion. It provides the necessary contextual conditioning for action generation without polluting the ground-truth physical history. Ultimately, this constant-memory paradigm fundamentally resolves the escalating peak memory and linearly increasing inference latency that traditionally plague autoregressive world models during long-horizon tasks.

\subsection{Speculative Asynchronous Inference}

\begin{figure}
    \centering
    \includegraphics[width=0.9\linewidth]{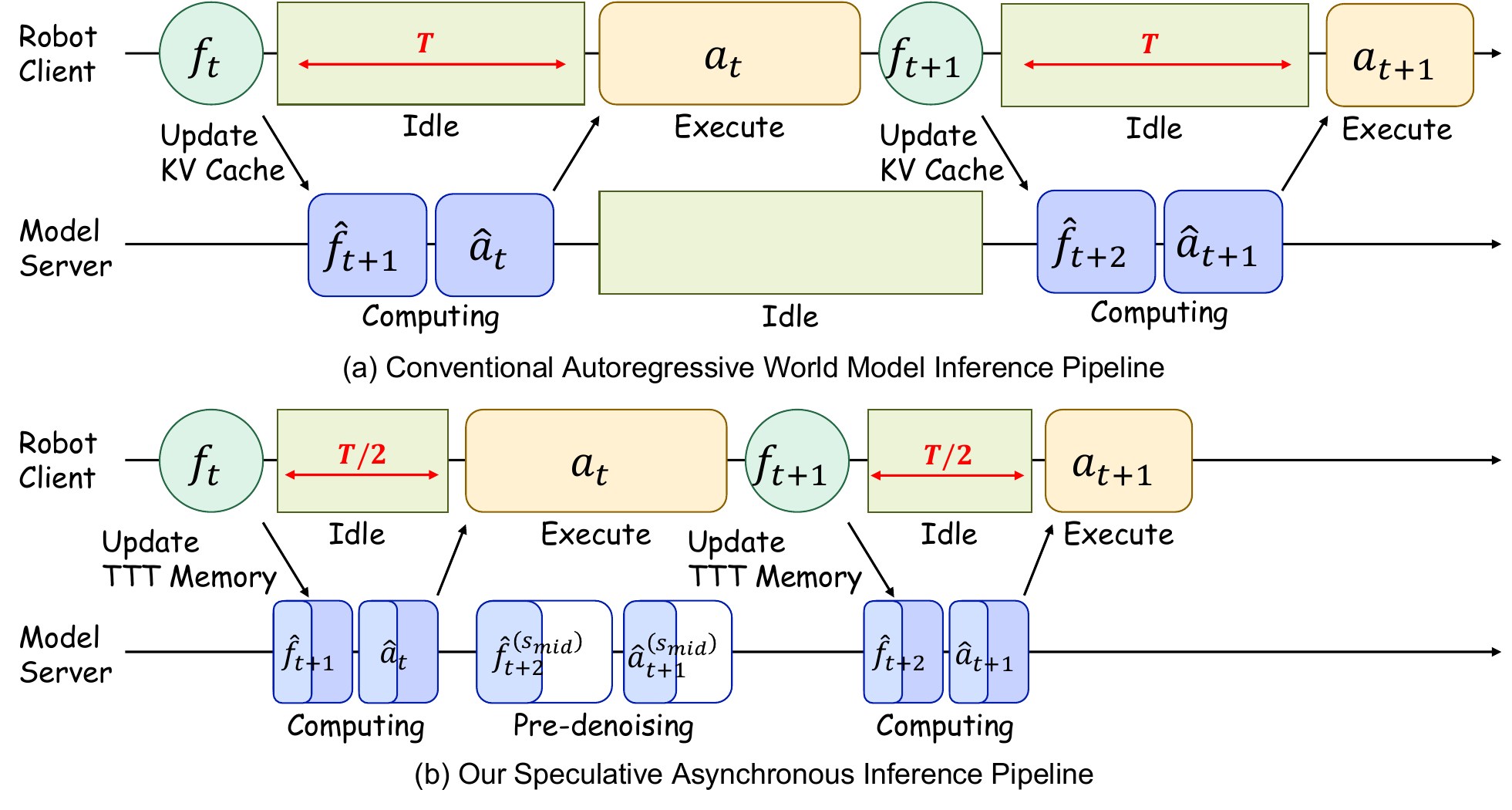}
    \caption{The Speculative Asynchronous Inference Pipeline. (a) Conventional autoregressive pipeline incurs high blocking latency by strictly waiting for the action execution and the true sensor observation $o_{t+1}$ / $f_{t+1}$ to arrive before next-step generation. (b) SAI leverages predicted future semantics $\hat{f}_{t+1}$ to proactively perform pre-denosing in the background. Upon observation concluding, new historical context are updated in TTT Memory modules, and only minimal fine-grained denoising is required to trigger the final predicted action $\hat{a}_{t+1}$. This effectively reduces the inference latency during deployment.}
    \label{fig:SAI_pipeline}
\end{figure}

While the TTT Memory module successfully bounds the per-step computational and memory overhead, the conventional "sense-compute-act" paradigm still enforces a strictly sequential execution pipeline. The policy must physically execute an action, wait for the movement to conclude, capture the next true observation $o_t$, and only then initiate the computationally heavy ODE integration. To further compress the overall deployment latency and break this sequential bottleneck, we propose the \textbf{Speculative Asynchronous Inference (SAI)} strategy. SAI leverages the forward-predictive causality of our model to deeply overlap neural computation with physical execution time, as shown in Fig.~\ref{fig:SAI_pipeline}.

\noindent\textbf{Phase 1: Speculative Pre-Denoising via Working Memory.} During the physical execution of action chunk $a_{t-1}$, the true future observation $o_t$ remains strictly inaccessible. However, CLWM has already anticipated the semantic state $\hat{f}_t$ during the preceding generation step. Instead of idling the GPU, SAI treats $\hat{f}_t$ as a surrogate observation. Leveraging our dual-state architecture, we construct a speculative context by updating the transient \textit{Working TTT Memory} with $\hat{f}_t$.Conditioned on this speculative hidden state, we proactively initiate the cascaded flow matching for step $t$. Specifically, the ODE solvers integrate the initial random noise from flow time $s=0$ up to an intermediate threshold $s = s_{\text{mid}}$ (where $0 < s_{\text{mid}} < 1$). This computationally intensive pre-denoising phase executes entirely in the background, perfectly masked by the duration of the robot's ongoing physical motion.

\noindent\textbf{Phase 2: Instantaneous Calibration via Long-Term Memory.} Upon the completion of the physical execution of $a_{t-1}$, the true sensor observation $o_t$ is instantly captured, yielding its exact DINOv3 semantic feature $f_t$. At this critical juncture, a rapid calibration occurs: the \textit{Long-Term TTT Memory} is officially anchored with the ground-truth $f_t$, and the speculative context previously driving the ODE solver is seamlessly swapped with the calibrated true memory state. The ODE integration then immediately resumes from $s = s_{\text{mid}}$ to complete the trajectory at $s = 1$. Because the model is only required to sequentially compute the remaining fine-grained denoising steps (the $1 - s_{\text{mid}}$ interval), the actual blocking latency experienced by the closed-loop control system is drastically minimized.

\noindent\textbf{Theoretical Synergy with History Augmentation.} Crucially, the mathematical stability of SAI is fundamentally guaranteed by the history augmentation strategy introduced in Eq.~\ref{eq:augment_history}. During Phase 1, the generative vector fields are inherently conditioned on an imperfect, speculative history. Because our MoT backbone was explicitly trained to deduce accurate flow velocities from noisy trajectories, it maintains highly robust directional gradients during the early flow steps $s \in [0, s_{\text{mid}}]$. This ensures that the pre-denoised representations remain securely bounded within the optimal transport path, allowing the Phase 2 calibration to effectively guide the final generation toward high-precision physical actions.By structurally decoupling algorithmic complexity from deployment latency, SAI unlocks high-frequency, reactive closed-loop control for complex embodied systems without sacrificing the expressive capacity of deep flow matching.
\section{EmbodiChain: Automating Robot Data Streaming via Generative Simulation}\label{sec:EmbodiChain}

Training World–Action Models (WAMs)\citep{li2025unified, zhu2025unified, bi2025motus, li2026causal, ye2026world} follows a data-driven paradigm. While their video-generation backbones benefit from large-scale online corpora~\citep{wan2025wan}, effective fine-tuning still relies on robot operation videos capturing rich spatio–temporal interactions among objects, robots, and language instructions.
Unlike language or vision models that can absorb internet-scale datasets, such robot data must be collected through physically grounded interactions within 3D environments. Generating such coherent and physically valid experiences requires costly simulation or real-world experiments, creating a mismatch between rapidly growing model capacity and the slower rate of experience production. In this regime, progress depends not on model size but on the efficiency of generating and consuming diverse interaction data.

\begin{wrapfigure}{r}{0.5\textwidth}
    \centering
    \vspace{-0.3in}
    \includegraphics[width=\linewidth]{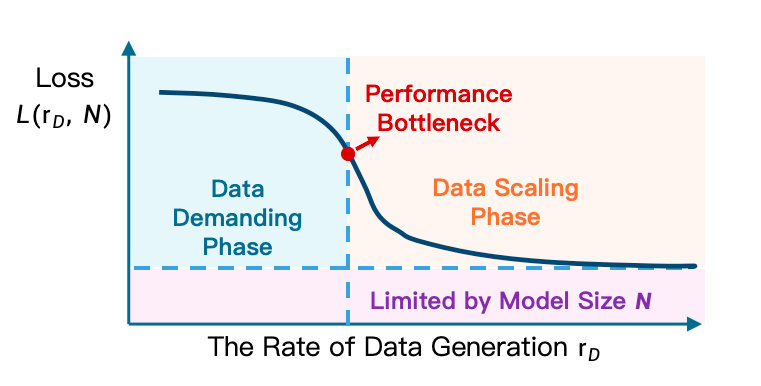}
    \vspace{-0.3in}
    \caption{Schematic illustration of the Efficiency Law: loss as a function of the rate of data generation.}\label{fig:efficiency-law}
    \vspace{-0.2in}
\end{wrapfigure}
A critical principle for overcoming this fundamental constraint is the establishment of the \textit{Efficiency Law of Embodied Intelligence}~\citep{GS-World}, as Fig.~\ref{fig:efficiency-law}, which posits that the effectiveness of embodied training depends primarily on maintaining a continuous flow of fresh, diverse, and physically valid experiences during learning. Sustaining this flow enables a regime of continual experience accumulation, where data generation and policy improvement co-evolve in real time, accelerating overall Sim2Real generalization.

Guided by this principle, we incorporate EmbodiChain~\citep{EmbodiChain} into our CLWM training framework.
EmbodiChain operationalizes the Efficiency Law through a closed-loop paradigm encompassing generative simulation, large-scale data expansion, and Sim2Real generalization, as detailed below.

\subsection{Generative Simulation for Robot Learning Environment} 

While traditional simulation systems~\citep{todorov2012mujoco,mittal2025isaaclab} offer rich interfaces that enable users to manually construct robotic environments, the number of environments available for training remains limited due to the substantial effort required for manual design and validation. This scarcity restricts their ability to capture the diverse distributional characteristics of real-world environments. EmbodiChain addresses this challenge of asset and scene diversity through a physics-aware two-stage generative process as follows: 

\begin{wrapfigure}{r}{0.5\textwidth}
    \centering
    \vspace{-0.1in}
    \includegraphics[width=\linewidth]{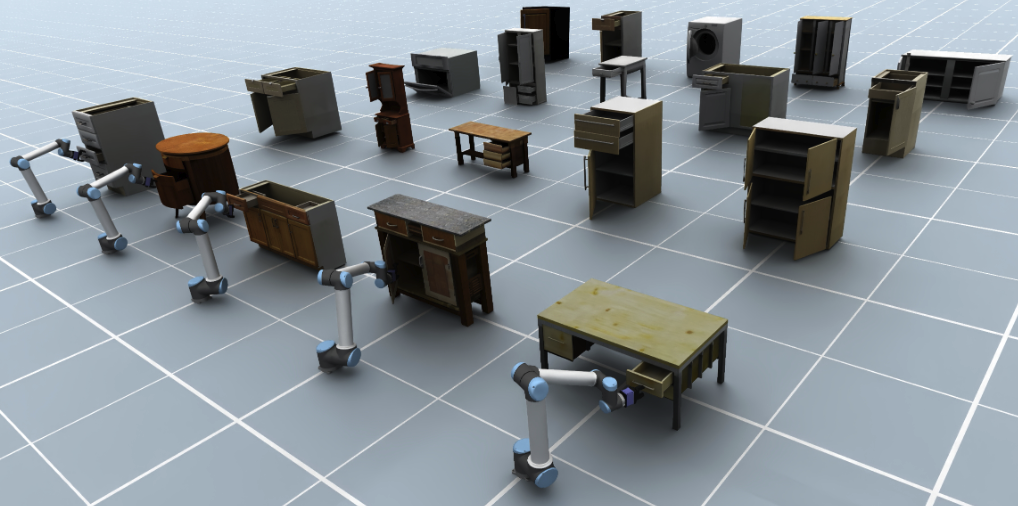}
    \vspace{-0.2in}
    \caption{Articulated 3D objects generated by predicting a part-decomposed structure, synthesizing  part geometry and appearance, and estimating articulation parameters for physics-based simulation~\citep{liu2026pact}.}\label{fig:PAct-demo}
    \vspace{-0.2in}
\end{wrapfigure}
\noindent\textit{1) Asset Generation and Optimization.}
A critical step in expanding environmental diversity is the generation of raw 3D meshes using generative models \citep{xiang2025trellis2}. However, these meshes often lack the geometric fidelity and physical realism required for high-quality simulation. EmbodiChain addresses this limitation through multi-objective optimization of each asset’s geometry, scale, and coordinate frame to ensure simulation compatibility. 
Fig.~\ref{fig:PAct-demo} illustrates an example.
We employ simulation-based validation to verify and refine key physical parameters, including mass distribution, friction coefficients, and collision properties. In addition, the system automatically computes interaction-critical attributes such as grasp poses and functional affordances. The refined assets are ultimately exported as simulation-ready Universal Scene Description (USD) files enriched with comprehensive physical and semantic metadata.


\noindent\textit{2) Scene Layout Synthesis.}
Building on the generated assets, EmbodiChain configures robot learning environments using scene generation methods \citep{hao2025mesatask} to produce an initial layout.
To bridge the gap between static scenes and functional workspaces, task-relevant interactive objects (foreground elements) are strategically positioned within the robot’s kinematically feasible region. The placement of background environmental assets is optimized through gradient-based refinement to eliminate inter-object penetrations and ensure a collision-free, physically plausible layout.

\begin{figure}[htbp]
    \centering
    \includegraphics[width=0.6\textwidth]{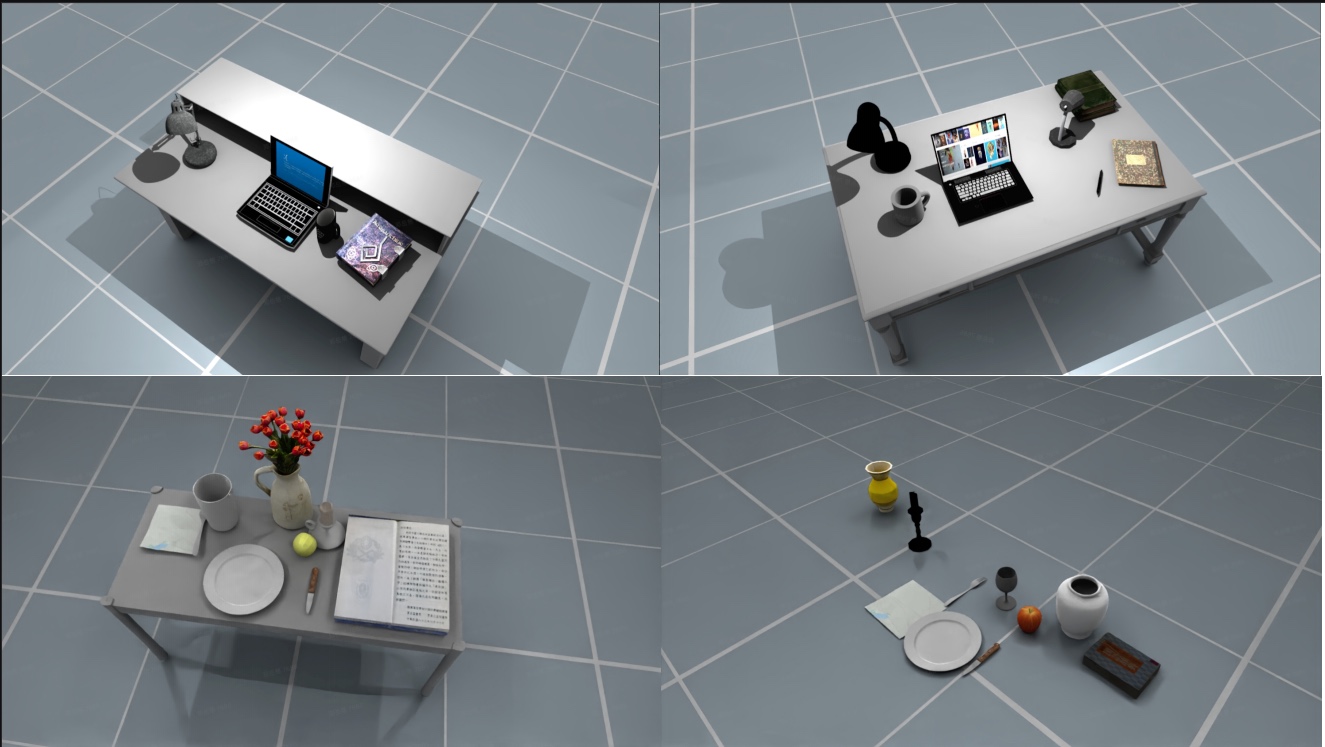}
    \caption{Example of a generated scene layout for robot learning environments, illustrating the placement of interactive objects and background assets to ensure a collision-free, physically plausible layout.}\label{fig:scene-gen}
\end{figure}


\subsection{Data Scaling via Domain Expansion} 
\label{sec:domain_expansion}
Building on the generated environments, EmbodiChain automatically generates and expands robot trajectories to address the limited coverage and lack of robustness in conventional embodied datasets. It introduces a unified data generation framework that jointly enhances functional diversity and enables failure-aware recovery.

\noindent\textit{1) Reachability-Aware Sampling.} A key factor limiting the diversity of robot motion is trajectory homogenization, which arises from human biases during teleoperation or from traditional motion planners that tend to converge to repetitive local optima. EmbodiChain mitigates this issue through a reachability-aware sampling strategy that promotes diversity within the task-relevant motion space rather than relying solely on raw joint configurations. We sample candidate robot states within the kinematically feasible workspace and select those that maximize dissimilarity across task-centric features such as end-effector approach direction, contact geometry, and interaction outcomes.
By prioritizing diversity in task-space representations while maintaining feasibility constraints, EmbodiChain generates trajectories that are both physically executable and functionally distinct, thereby enriching the dataset with a wide range of manipulation strategies.

\begin{figure}[H]
    \centering
    \includegraphics[width=0.75\textwidth]{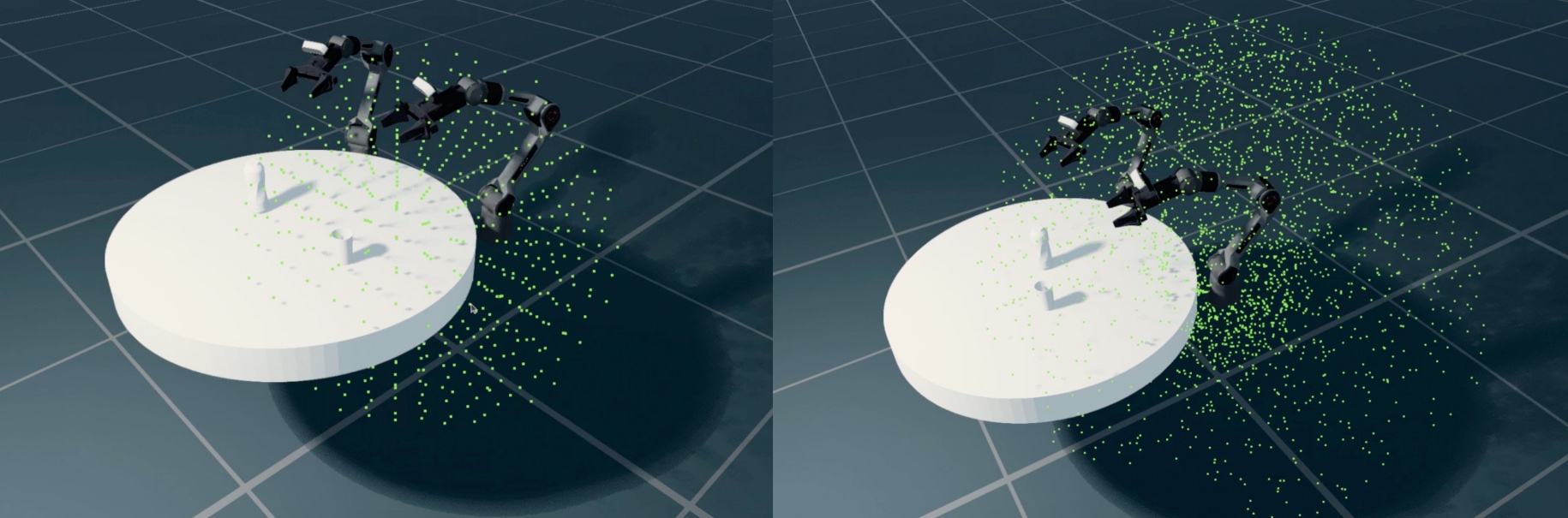}
    \caption{Robot workspace visualization.}
    \label{fig:robot-ws}
\end{figure}

\noindent \textit{2) Closed-loop Error Recovery.} To enhance the efficiency and robustness of the diversity-driven sampling, EmbodiChain incorporates a closed-loop error recovery mechanism.
When failures occur (e.g., object slippage, misaligned grasps, or boundary violations), a reactive replanning module generates corrective motion trajectories that steer the system back toward task completion. These recovery sequences are relabeled and reintegrated into the dataset, providing structured supervision on how to act under perturbed or unstable conditions.
By transforming execution failures into informative training signals, this process augments the data distribution with recovery-critical behaviors that are largely absent in conventional datasets.


\noindent\textit{3) Visual Augmentation.}
To bridge the visual gap between simulated and real-world observations, EmbodiChain employs a streaming-based parametric synthesis mechanism designed for generative world modeling. Rather than applying static post-processing, this module dynamically samples from a high-dimensional distribution of environmental factors, including lighting temperatures, surface BRDF properties, and continuous sensor drift, during the on-the-fly data generation process. Unlike traditional augmentation techniques that risk breaking temporal causal links, EmbodiChain enforces temporal consistency constraints by evolving these parameters through smooth stochastic processes. This enables the world model to decouple invariant transition dynamics from transient visual appearances. By enriching the visual domain in a continuous, latent-conditioned manner, the system forces the model to capture generalizable geometric and semantic world states rather than overfitting to simulation-specific rendering artifacts. This parametric diversity provides a rigorous foundation for Sim2Real transfer, allowing the generative world model to synthesize and predict across the wide distribution of real-world visual conditions.

\noindent\textit{4) Physics-Grounded Generation.}
An important prerequisite for effective data scaling is maintaining physical validity. Building upon the rigorously optimized assets and collision-free scene layouts generated in the previous stage, EmbodiChain ensures that all expanded domains strictly adhere to the principles of classical mechanics. Rather than unconstrained randomization, the scaling process preserves the structural integrity of multi-body articulations and the consistency of physical parameters, such as mass distributions and dynamic friction, across all varied scenarios. By grounding this vast diversity in physical reality, EmbodiChain guarantees that the learned behaviors remain physically executable.

\subsection{Efficient Scaling via Online Data Streaming}

\noindent\textbf{The Physical Bottleneck of Embodied Scaling.} While large vision and language models benefit from passively absorbing internet-scale corpora, embodied AI is intrinsically constrained by the physical generation rate of valid trajectories. Thus, traditional scaling strategies that merely enlarge static datasets fail to provide the continuous, diverse interactions required for generalized robust control.

\noindent\textbf{Efficiency Law and Experience Throughput.} To address this \emph{bottleneck of physicality}, we draw upon the \emph{Efficiency Law} of embodied intelligence \cite{GS-World}, which emphasizes maintaining a high, dynamic data generation rate over total static dataset size. We measure this via \textbf{Experience Throughput} ($\mathcal{E}$), the volume of unique state-action pairs ingested per training iteration. For fixed compute ($C$) and parameters ($P$), intelligence ($\mathcal{I}$) scales effectively only when $\mathcal{E}$ exceeds a critical threshold $\tau(C,P)$. Therefore, our optimization target shifts to maximizing \textbf{dynamic experience density}: prioritizing high-throughput, novel experiences to ensure highly informative gradient updates and rapid policy improvement.

\noindent\textbf{Asynchronous Streaming Architecture.} EmbodiChain introduces \emph{Online Data Streaming} (ODS), a storage-less paradigm that continuously synthesizes and directly injects fresh trajectories into the optimizer. Exploiting a heterogeneous shared-memory pipeline, simulation and generation workers asynchronously write to a lock-free circular buffer in CPU and GPU VRAM. Learner workers consume batches directly via zero-copy exchange, eliminating costly serialization and I/O bottlenecks. A bounded reuse mechanism amortizes generation costs while strictly preserving sample novelty.

\noindent\textbf{Infinite Diversity for Superior Generalization.} Conventional offline training often suffers from overfitting due to the finite cardinality of static datasets. In contrast, ODS overlaps the generation of diverse, failure-aware trajectories (Section~\ref{sec:domain_expansion}) directly with model updates, enabling \emph{unbounded data diversity at near-zero marginal time cost}. This continuous influx of out-of-distribution variations prevents the model from memorizing static environments, acting as the critical driver for learning robust, generalized strategies.

\noindent\textbf{A Unified Learning Paradigm.} ODS conceptually unifies traditionally distinct paradigms. It preserves the reactive nature of \textbf{Online RL} through a continuous data flow, while its high-speed VRAM buffering retains the computational stability of \textbf{Offline RL}. Ultimately, ODS provides an optimized framework to apply stable \textbf{Supervised Learning} objectives on a virtually infinite, on-the-fly generated data pool.

\section{Experiments}

\subsection{Dataset Curation}

\subsubsection{Pretraining Data}

To achieve strong generalization across diverse manipulation tasks, our pretraining stage utilizes a large-scale corpus aggregated from open-source robot manipulation datasets, mainly including RoboMind, Agibot World Beta and InternData-A1 datasets. All datasets undergo rigorous preprocessing to ensure consistency in data formatting and annotation quality. This massive offline dataset enables the model to learn robust causal world dynamics and general-purpose visual representations prior to task-specific finetuning.
Specifically, for video inputs, we utilize the DINOv3 base model~\citep{simeoni2025dinov3} to extract latent feature maps from the raw video frames. For action representation, we follow the LingBot-VA approach~\citep{li2026causal} to unify diverse action representations into a standardized format. The total action dimensionality for dual-arm systems comprises 7-DoF end-effector poses, 7 joint positions, and 1 gripper state per arm, yielding a highly compact 30-dimensional continuous action space: $(7 + 7 + 1) \times 2$.

\subsubsection{Post-training Data}
For post-training and task-specific adaptation, we entirely eschew the manual collection of real-world or downstream demonstrations. Instead, we rely exclusively on data generated through our EmbodiChain framework. By continuously synthesizing customized, physics-grounded simulated trajectories, we force the model to bridge the sim-to-real gap autonomously without requiring costly human-in-the-loop data collection.

\subsection{Implementation \& Training Details}

\noindent\textbf{Model Architecture Configurations.} Our Causal Latent World Model (CLWM) utilizes the pre-trained DINOv3 base model as a semantic feature extractor with a patch size of $P=16$. The generative process is driven by a highly parameter-efficient Mixture of Transformers (MoT) backbone initialized from Wan2.2-5B, augmented with domain-specific projection layers. Action decoding employs an action chunk size of $\tau=16$. To overcome sequence-scaling barriers, our Dual-State TTT Memory is instantiated via a TTT-MLP architecture featuring a $4\times$ expansion factor and GELU activations. The gating vector $\alpha$ is initialized to $0.1$ to guarantee stable fine-tuning dynamics.

\noindent\textbf{Training \& Hyperparameters.} Pretraining utilizes the AdamW optimizer with a base learning rate of $1 \times 10^{-4}$ and a global batch size of $128$. The objective is optimized over about $20$ epochs on the aggregated manipulation datasets. Furthermore, to enhance the robustness of the Action Model against imperfect visual histories, we enforce the history augmentation strategy by injecting Gaussian noise with probability $p=0.5$ at varying scales $s_{\text{aug}} \in [0.5, 1]$.

\noindent\textbf{Compute Infrastructure.} All large-scale offline pretraining and subsequent online adaptations are distributed and executed across a computing cluster composing $64$ NVIDIA H100 GPUs. The entire training demands a continuous computational duration of approximately $20$ days.

\subsection{Main Results}

We comprehensively evaluate our approach across a rigorous simulated benchmark and physical robot deployments. Our experiments are designed to validate three core dimensions: (1) complex manipulation and bimanual coordination capabilities, (2) the architectural efficiency of our Dual-State Memory and SAI, and (3) the data scaling efficacy of the EmbodiChain framework.

\subsubsection{Simulation Results on RoboTwin}

We benchmark CLWM in simulation using the challenging RoboTwin environment, which demands high-precision dual-arm coordination and complex object-centric interactions. By embedding strong structural priors through the causal world modeling, CLWM demonstrates unprecedented policy robustness. As detailed in Table~\ref{tab:robotwin_results}, our method establishes a new state-of-the-art, consistently outperforming all established baseline policies across the vast majority of tasks, achieving an average success rate of $94.00\%$.

\FloatBarrier
\begin{table}[htbp]
    \centering
    \resizebox{0.7\linewidth}{!}{
    \begin{tabular}{lccccc}
        \toprule
        \textbf{Simulation Task} & \textbf{$\pi_{0.5}$} & \textbf{X-VLA} & \textbf{Motus} & \textbf{LingBot-VA} & \textbf{Ours} \\
        \midrule
        \textit{Adjust Bottle} & 99\% & 99\% & 93\% & 94\% & \textbf{100\%}  \\
        \textit{Beat Block Hammer} & 93\% & 88\% & 88\% & \textbf{98\%} & \textbf{98\%}  \\
        \textit{Blocks Ranking RGB} & 85\% & 83\% & 97\% & \textbf{98\%} & \textbf{98\%}  \\
        \textit{Blocks Ranking Size} & 26\% & 74\% & 63\% & 96\% & \textbf{97\%}  \\
        \textit{Click Alarmclock} & 89\% & 99\% & \textbf{100\%} & \textbf{100\%} & \textbf{100\%}  \\
        \textit{Click Bell} & 66\% & \textbf{100\%} & \textbf{100\%} & \textbf{100\%} & \textbf{100\%}  \\
        \textit{Dump Bin Bigbin} & \textbf{97\%} & 77\% & 91\% & 96\% & 96\%  \\
        \textit{Grab Roller} & \textbf{100\%} & \textbf{100\%} & \textbf{100\%} & \textbf{100\%} & \textbf{100\%}  \\
        \textit{Handover Block} & 57\% & 37\% & 73\% & 78\% & \textbf{80\%}  \\
        \textit{Handover Mic} & 97\% & 0\% & 63\% & 96\% & \textbf{97\%}  \\
        \textit{Hanging Mug} & 17\% & 27\% & 38\% & 28\% & \textbf{40\%}  \\
        \textit{Lift Pot} & 85\% & \textbf{100\%} & 99\% & 99\% & \textbf{100\%}  \\
        \textit{Move Can Pot} & 55\% & 86\% & 74\% & \textbf{97\%} & 95\%  \\
        \textit{Move Pillbottle Pad} & 61\% & 71\% & 96\% & \textbf{99\%} & \textbf{99\%}  \\
        \textit{Move Playingcard Away} & 84\% & 98\% & 96\% & \textbf{99\%} & \textbf{99\%}  \\
        \textit{Move Stapler Pad} & 42\% & 73\% & 85\% & 79\% & \textbf{86\%}  \\
        \textit{Open Laptop} & 96\% & \textbf{100\%} & 91\% & 94\% & \textbf{100\%}  \\
        \textit{Open Microwave} & 77\% & 71\% & 91\% & 86\% & \textbf{93\%}  \\
        \textit{Pick Diverse Bottles} & 71\% & 36\% & \textbf{91\%} & 82\% & 85\%  \\
        \textit{Pick Dual Bottles} & 63\% & 36\% & 90\% & 99\% & \textbf{100\%}  \\
        \textit{Place A2B Left} & 82\% & 49\% & 79\% & 93\% & \textbf{95\%}  \\
        \textit{Place A2B Right} & 84\% & 36\% & 87\% & \textbf{95\%} & \textbf{95\%}  \\
        \textit{Place Bread Basket} & 64\% & 71\% & 94\% & 95\% & \textbf{96\%}  \\
        \textit{Place Bread Skillet} & 66\% & 67\% & 83\% & 90\% & \textbf{93\%}  \\
        \textit{Place Burger Fries} & 87\% & 94\% & \textbf{98\%} & 95\% & 96\%  \\
        \textit{Place Can Basket} & 62\% & 52\% & 76\% & 84\% & \textbf{86\%}  \\
        \textit{Place Cans Plasticbox} & 84\% & 98\% & 94\% & \textbf{99\%} & \textbf{99\%}  \\
        \textit{Place Container Plate} & 95\% & 95\% & 99\% & 97\% & \textbf{99\%}  \\
        \textit{Place Dual Shoes} & 75\% & 88\% & 87\% & 89\% & \textbf{91\%}  \\
        \textit{Place Empty Cup} & 99\% & 98\% & 98\% & \textbf{100\%} & \textbf{100\%}  \\
        \textit{Place Fan} & 85\% & 75\% & 87\% & 93\% & \textbf{95\%}  \\
        \textit{Place Mouse Pad} & 39\% & 70\% & 68\% & 96\% & \textbf{98\%}  \\
        \textit{Place Object Basket} & 76\% & 39\% & 87\% & 88\% & \textbf{89\%}  \\
        \textit{Place Object Scale} & 80\% & 74\% & 85\% & 95\% & \textbf{97\%}  \\
        \textit{Place Object Stand} & 85\% & 88\% & 97\% & 96\% & \textbf{98\%}  \\
        \textit{Place Phone Stand} & 81\% & 87\% & 86\% & 97\% & \textbf{99\%}  \\
        \textit{Place Shoe} & 93\% & 95\% & 97\% & \textbf{98\%} & \textbf{98\%}  \\
        \textit{Press Stapler} & 83\% & 98\% & 98\% & 82\% & \textbf{99\%}  \\
        \textit{Put Bottles Dustbin} & 79\% & 77\% & 79\% & 91\% & \textbf{93\%}  \\
        \textit{Put Object Cabinet} & 79\% & 48\% & 71\% & 87\% & \textbf{88\%}  \\
        \textit{Rotate QRcode} & 87\% & 33\% & 73\% & 91\% & \textbf{94\%}  \\
        \textit{Scan Object} & 65\% & 36\% & 66\% & 91\% & \textbf{92\%}  \\
        \textit{Shake Bottle Horizontally} & 99\% & \textbf{100\%} & 98\% & 99\% & \textbf{100\%}  \\
        \textit{Shake Bottle} & 97\% & \textbf{100\%} & 97\% & 97\% & \textbf{100\%}  \\
        \textit{Stack Blocks Three} & 76\% & 10\% & 95\% & 98\% & \textbf{100\%}  \\
        \textit{Stack Blocks Two} & \textbf{100\%} & 87\% & 98\% & 98\% & \textbf{100\%}  \\
        \textit{Stack Bowls Three} & 71\% & 86\% & 87\% & 83\% & \textbf{88\%}  \\
        \textit{Stack Bowls Two} & 96\% & 93\% & 98\% & 94\% & \textbf{97\%}  \\
        \textit{Stamp Seal} & 55\% & 82\% & 92\% & 96\% & \textbf{97\%}  \\
        \textit{Turn Switch} & 54\% & 61\% & 78\% & 44\% & \textbf{65\%}  \\
        \midrule
        \textbf{Average (\%)} & 76.76\% & 72.84\% & 87.02\% & 91.55\% & \textbf{94.00\%} \\
        \bottomrule
    \end{tabular}
    }
    \caption{Success rates on the RoboTwin benchmark across various tasks. Our method (CLWM) demonstrates superior performance compared to previous baseline policies.}
    \label{tab:robotwin_results}
\end{table}


\noindent\textbf{Training Specifications.} For the extensive evaluations on RoboTwin, all models are fine-tuned on a robust dataset comprising $25,000$ synthetic trajectories. The optimization process is scaled to $40$k iterations with the learning rate of $1\times 10^{-5}$ to effectively adapt the semantic priors acquired during pre-training into robust, task-specific control policies.

\subsubsection{Efficiency Analysis}
To validate the architectural claims proposed in this work, we evaluate the deployment efficiency of CLWM during long-horizon physical inferences.

\noindent\textbf{Constant Memory via Dual-State TTT.} We compared the peak GPU memory footprint of our Dual-State TTT Memory against a standard Transformer KV-Cache baseline during an extended 2,000-step manipulation episode. While the traditional KV-Cache exhibits a strict linear memory explosion $\mathcal{O}(T)$, our TTT mechanism maintains a perfectly flat, $\mathcal{O}(1)$ constant memory footprint throughout the entire episode length, proving its viability for unbounded continuous deployment.

\noindent\textbf{Decreasing Latency via SAI.} To quantify the benefits of Speculative Asynchronous Inference (SAI), we profiled the end-to-end blocking latency (the time the physical robot waits for neural computation). By overlapping the computationally heavy ODE pre-denoising ($s=0 \to s_{\text{mid}}$) with the robot's physical execution, SAI drastically reduces the blocking latency by about $\mathbf{50\%}$ compared to a strictly sequential autoregressive pipeline. This transforms the effective closed-loop control frequency from a sluggish baseline to a highly reactive high-frequency regime, critical for real-world dynamic perturbations.

\subsubsection{EmbodiChain Experiments}
To evaluate the capabilities of the EmbodiChain framework, we focus our ablation studies on three representative manipulation tasks: \textit{Hanging Mug}, \textit{Turn Switch}, and \textit{Stack Bowls}. These tasks are selected to reflect varying degrees of geometric complexity, precision requirements, and multi-step coordination.

\noindent\textbf{Ablation Study on Domain Expansion}
We evaluate the impact of EmbodiChain's domain expansion techniques by systematically isolating the contributions of visual augmentation, physics-grounded generation, and reachability-aware sampling. We establish a baseline using only \textit{Spatial Randomization} (randomizing object positions and orientations) and progressively introduce the subsequent modules. For a fair comparison, all configurations are trained on an identical budget composed of $2,000$ synthesized trajectories per task ($6,000$ trajectories in total). To rigorously assess robustness, evaluations are conducted under both In-Distribution (ID) and Out-of-Distribution (OOD) conditions, where OOD scenarios introduce unseen objects, novel textures, extreme lighting variations, and unfamiliar spatial layouts.

As shown in Tab.~\ref{tab:ablation_domain_expansion}, while spatial randomization and visual augmentation achieve reasonable performance on ID tasks, they suffer a severe performance drop in OOD settings due to overfitting to spurious visual correlations. Integrating physics-grounded generation bridges this gap, allowing the policy to capture the underlying causal dynamics of the physical interaction. Ultimately, the addition of reachability-aware sampling (our full pipeline) yields the highest success rates across both ID and OOD configurations by preventing trajectory homogenization and teaching the policy to recover from diverse perturbations and initial states.

\begin{table}[H]
    \centering
    \resizebox{0.8\linewidth}{!}{
    \begin{tabular}{lcc}
        \toprule
        \textbf{Configuration} & \textbf{ID Success (\%)} & \textbf{OOD Success (\%)} \\
        \midrule
        Baseline (Spatial Randomization Only) & 64\% & 25\% \\
        + Visual Augmentation & 75\% & 42\% \\
        + Physics-grounded Generation & 81\% & 56\% \\
        + Reachability-aware Sampling (Full) & \textbf{95\%} & \textbf{82\%} \\
        \bottomrule
    \end{tabular}
    }
    \caption{Ablation study on EmbodiChain domain expansion techniques. Average success rates (\%) are reported under In-Distribution (ID) and Out-of-Distribution (OOD) conditions (evaluating over unseen objects, textures, lighting, and spatial layouts).}
    \label{tab:ablation_domain_expansion}
\end{table}

\noindent\textbf{Ablation Study on Online Data Streaming}
To validate the efficacy of our Online Data Streaming (ODS) pipeline and the underlying Efficiency Law of Embodied Intelligence, we compare ODS against a traditional static dataset training paradigm on these three manipulation tasks. 

For a fair comparison, all configurations share a training budget of 5,000 iterations (batch size 64). The static baseline trains on a fixed dataset of 1,500 demonstrations (500 per task), dictating an expected sampling frequency of approximately 213 times per trajectory ($\frac{64 \times 5000}{1500} \approx 213$). In contrast, ODS continuously synthesizes and streams new trajectories into a shared memory buffer. We ablate the maximum replay frequency of the streamed data, evicting trajectories after they are sampled 213, 50, or 10 times ($\text{ODS}_{\text{sample } 213}$, $\text{ODS}_{\text{sample } 50}$, and $\text{ODS}_{\text{sample } 10}$). A lower replay limit enforces faster data turnover, injecting a higher throughput of novel experiences into the optimization process.

Tab.~\ref{tab:ablation_ods} highlights the transformative advantage of our approach. Notably, when the ODS replay bound is artificially inflated to match the baseline's sample count ($\text{ODS}_{\text{sample } 213}$), performance degenerates to the static regime (e.g., $60\%$ vs. $62\%$ on \textit{Hanging Mug}), as the buffer stagnates and suffers from trajectory homogenization. 

However, as the replay factor decreases, performance improves monotonically. Limiting trajectory reuse to 50 times ($\text{ODS}_{\text{sample } 50}$) yields up to a $30\%$ absolute improvement over the static baseline. Further compressing the replay bound to 10 ($\text{ODS}_{\text{sample } 10}$) maximizes dynamic experience density, pushing success rates to near-perfect levels ($96\%$--$98\%$). These findings substantiate the Efficiency Law: scaling the continuous throughput of fresh physical interactions, rather than static dataset size, is the fundamental driver for policy robustness and generalization.

\begin{table}[H]
    \centering
    \resizebox{0.8\linewidth}{!}{
    \begin{tabular}{lccc}
        \toprule
        \textbf{Training Configuration} & \textbf{Hanging Mug} & \textbf{Turn Switch} & \textbf{Stack Bowls} \\
        \midrule
        Static Baseline (1,500 demos) & 62\% & 85\% & 88\% \\
        \midrule
        $\text{ODS}_{\text{sample } 213}$ & 60\% & 84\% & 85\% \\
        $\text{ODS}_{\text{sample } 50}$  & 92\% & 92\% & 96\% \\
        $\text{ODS}_{\text{sample } 10}$  & \textbf{96\%} & \textbf{98\%} & \textbf{98\%} \\
        \bottomrule
    \end{tabular}
    }
    \caption{Ablation studies on Online Data Streaming (ODS) vs. static dataset training. All configurations use 5,000 training iterations and a batch size of 64. A lower replay bound in ODS indicates higher throughput of novel experiences.}
    \label{tab:ablation_ods}
\end{table}

\subsubsection{Real-world Deployment}
To validate real-world effectiveness, we deploy our model on  Agilex CobotMagic bimanual platform and test it across four challenging everyday manipulation tasks demanding precision and multi-step reasoning: \textit{Dual-Arm Water Pouring}, \textit{Table Rearrangement}, \textit{Items Hand-Over and Place}, and \textit{Pan Open and Place}. 
Importantly, our model (CLWM) successfully performs these real-world tasks despite being trained \textit{exclusively} on simulation data. This sim-to-real transfer is enabled by the online data streaming and domain expansion techniques introduced via the EmbodiChain framework detailed in Sec.~\ref{sec:EmbodiChain}.

\noindent\textbf{Training Specifications:} For a fair evaluation, we standardize the training data budget across our method and the baselines. For the fully sim-to-real methods (Sim2Real-VLA and our CLWM), the models are trained purely in simulation using the EmbodiChain domain expansion and Online Data Streaming (ODS) pipelines with identical synthetic training budgets. In contrast, the baseline real-world policies ($\pi_0$ and GR00T N1.5) are finetuned using exactly 50 real-world expert demonstrations per task, as they lack an automated sim-to-real synthetic data generation mechanism. 
As highlighted in Tab.~\ref{tab:real_world}, our unified model exhibits strong zero-shot capabilities, achieving reliable success rates across all four tasks despite real-world sensory noise and lack of physical demonstration data.

\begin{table}[H]
    \centering
    \resizebox{\linewidth}{!}{
    \begin{tabular}{lcccc}
        \toprule
        \textbf{Methods} & \textbf{Dual-Arm Water Pouring} & \textbf{Table Rearrangement} & \textbf{Items Hand-Over and Place} & \textbf{Pan Open and Place} \\
        \midrule
        $\pi_0$ & 25\% & 20\% & 20\% & 5\% \\
        GR00T N1.5 & 35\% & 20\% & 15\% & 5\% \\
        Sim2Real-VLA & 80\% & 80\% & 40\% & 35\% \\
        CLWM (Ours) & \bf 95\% & \bf 90\% & \bf 80\% & \bf 65\% \\
        \bottomrule
    \end{tabular}
    }
    \caption{Real-world deployment success rates across four challenging manipulation tasks, evaluated zero-shot using only simulation training data. Results are reported as success fraction $\pm$ standard error.}
    \label{tab:real_world}
\end{table}

\section{Conclusion}

In this work, we presented the Causal Latent World Model (CLWM), an autoregressive world-action framework that addresses the critical representational, memory, and latency barriers in continuous robot deployment. By employing DINOv3 latent features as generative targets instead of raw pixels, CLWM effectively disentangles core interaction semantics from redundant visual texture, laying the foundation for robust domain generalization. Furthermore, by incorporating a Dual-State TTT Memory and Speculative Asynchronous Inference (SAI), CLWM achieves constant $\mathcal{O}(1)$ memory scaling and drastically decreases generation latency. Recognizing that robust physical policies demand massive, diverse interactions, we further introduced EmbodiChain to establish the \textit{Efficiency Law}. Through physics-grounded generative simulation and Online Data Streaming, EmbodiChain continuously injects novel, error-recovering trajectories into the optimization process, entirely bypassing the bottleneck of static dataset collection. Our comprehensive evaluations highlight the transformative potential of this unified system. CLWM achieves state-of-the-art performance in complex dual-arm simulated tasks and, crucially fueled by its latent semantic architecture and online data pipeline, demonstrates exceptional zero-shot sim-to-real transfer capabilities on physical hardware, outperforming baselines explicitly finetuned with real-world human data.



\bibliographystyle{plainnat}

\bibliography{main}

\end{document}